\pdfoutput=1

\documentclass[11pt]{article}

\usepackage[preprint]{acl}

\usepackage{times}
\usepackage{latexsym}

\usepackage[T1]{fontenc}

\usepackage[utf8]{inputenc}

\usepackage{microtype}

\usepackage{inconsolata}

\usepackage{graphicx}

\usepackage{amsmath}
\usepackage{amssymb}
\usepackage{mathtools}
\usepackage{amsthm}
\usepackage{amsfonts}
\usepackage{multirow}
\usepackage{booktabs}
\usepackage{makecell}
\usepackage[most]{tcolorbox}
\usepackage{algorithm}
\usepackage{algorithmic}

\newcommand{\ModelName}{{EmoFSM}} 
\newcommand{\vanillaFT}{{SFT}}

\title{EmoFSM: A Finite State Machine for Emotional Support Conversation}

\author{
\textbf{Yue Zhao},
 \textbf{Qingqing Gu},
 \textbf{Xiaoyu Wang},
 \textbf{Teng Chen},
 \textbf{Zhonglin Jiang},
 \textbf{Yong Chen},\\
 \textbf{Hongyan Li},
 \textbf{Luo Ji} 
\\
\\
  \textsuperscript{1} Geely AI Lab
\\
  \small{
    \textbf{Correspondence:} \href{mailto:email@domain}{Luo.Ji1@Geely.com}
  }
}

\begin{document}

\maketitle

\begin{abstract}
Emotional support conversation (ESC) aims to alleviate people's emotional distress through effective conversations. Although large language models (LLMs) have made remarkable progress in ESC, most of these studies may not define the diagram from a state-model perspective, thereby providing a suboptimal solution for long-term satisfaction. To address such an issue, we leverage the Finite State Machine (FSM) on LLMs, and propose a framework called {\ModelName}. Our framework allows a single LLM to bootstrap the planning during ESC, and self-reason the seeker's emotion, support strategy, and the final response upon each conversation turn. Substantial experiments in ESC datasets suggest that {\ModelName} outperforms many baselines, including direct inference, self-fine, chain of thought, finetuning, and externally supported methods, even those with many more parameters.
\end{abstract}

\section{Introduction}

Emotional support conversation (ESC) is a seeker-support dialogue scenario that aims to mitigate the seeker's emotional stress and personal challenges in real life \cite{heaney2008social,langford1997social}. Effective emotional support is related to relational, psychological, and physical theories \cite{rains2020support}, and has also been widely studied by artificial intelligence techniques \cite{liu2021ESconv,cheng-etal-2022-improving}. With the remarkable advances of Large Language Models (LLMs) in NLP tasks, LLM-based methods have achieved impressive results in ESC research \cite{zheng-etal-2023-augesc,kang-etal-2024-large}. However, most existing LLM-based approaches tend to provide a myopic solution to the help seeker, without incorporating forward-looking strategic planning. For instance, \citet{liu2021ESconv} proposes that a full ESC process typically includes three stages (Exploration $\rightarrow$ Comforting $\rightarrow$ Action), yet an LLM may fail to anticipate and guide these stage transitions, leading to pronounced strategy bias \cite{kang-etal-2024-large}.

\begin{figure}[htbp!]
\centering
  \includegraphics[width=0.99\linewidth]{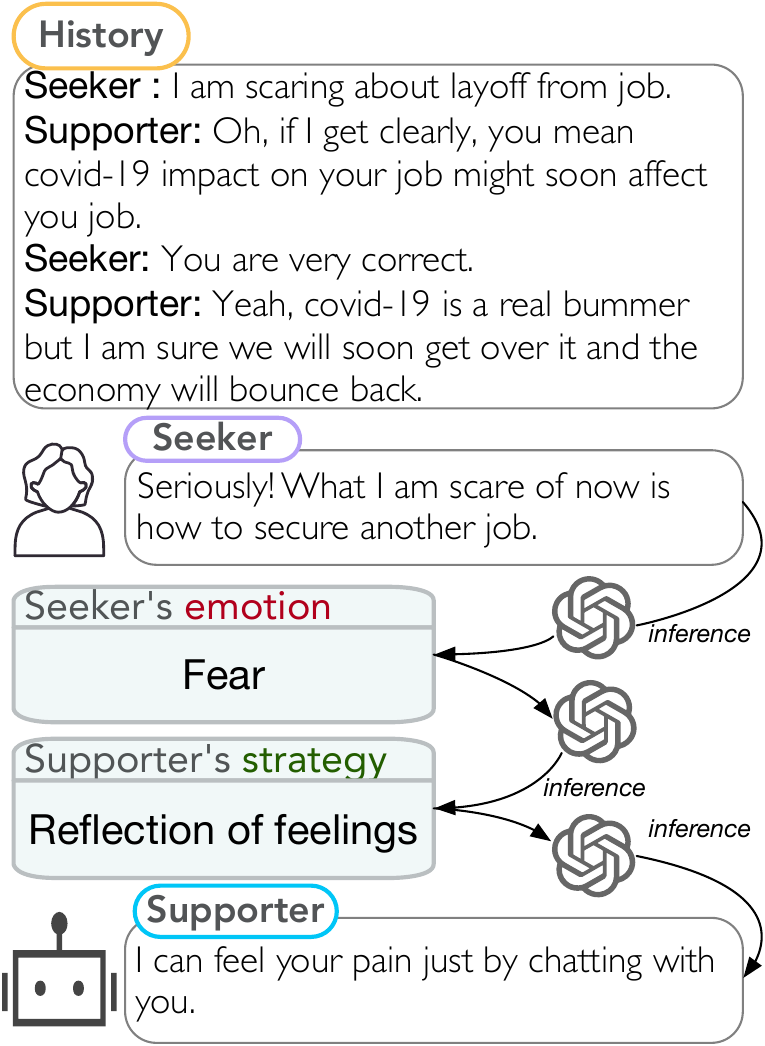} 
  \caption {The paradigm of {\ModelName} on emotional support conversations.}
  \label{fig:paradigm}
\end{figure}

Motivated by Finite State Machines (FSMs) to question-answering \cite{wang2024fsmfinitestatemachine,wang2024sgfsm} and reasoning tasks \cite{liu2024SMoT}, we argue that ESC tasks can also be modeled as a finite state diagram, allowing FSMs to alleviate the challenges outlined above. Given the seeker's current utterance and the dialogue history, the LLM is prompted to recognize the seeker's present emotion, select an appropriate support strategy, and generate its response. Once the current conversation turn ends, the system updates the history by appending the emotion, strategy, and utterances for that turn and then transitions back to the initial state. 


\begin{figure*}[t!]
\centering
  \includegraphics[width=0.8\linewidth]{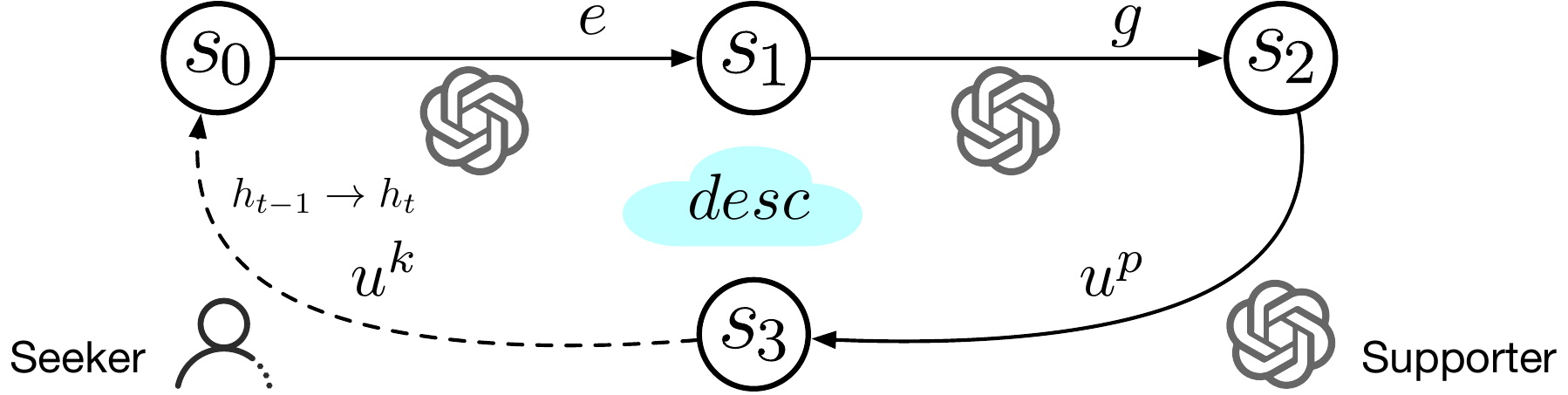} 
  \caption {The {\ModelName} framework, traversing four states of conversations.}
  \label{fig:framework}
\end{figure*}

In this paper, we propose a framework called \textbf{emo}tional \textbf{Fi}nite \textbf{S}tate \textbf{M}achine (\textbf{\ModelName}), which applies FSM to emotional support conversations, as shown in Figure \ref{fig:paradigm}. Upon each conversation turn, we allow the LLM to self-generate the emotion, strategy, and final utterance. We employ ESConv \cite{liu2021ESconv}, an annotated multi-turn ESC dataset to fine-tune the LLM. The strategic thinking paradigm is bootstrapped during the training, and the various state transition tasks are jointly optimized through this multitask learning framework. Based on comprehensive automatic and human evaluation, we find that our {\ModelName} outperforms different baselines, including direct inference, prompt-based approaches, standard fine-tuning, methods that integrate external knowledge, as well as hybrids of these techniques. Our key contributions are as follows:

\begin{itemize}

\item We propose that ESC tasks can be solved by FSM, and define the states as the combination of utterances, seeker's emotion, and supporter's strategy.
\item Our methodology utilizes LLM in a multi-hop paradigm, which triggers the strategic bootstrapping capability that can surpass the golden annotations.
\item Thorough experiments verify that our methodology outperforms different prompt, fine-tuning, and external knowledge-based baselines, either on similar or much larger sizes.

\end{itemize}

\section{Methodology}

This section introduces the methodology details. The finite state diagram is shown in Figure \ref{fig:framework}. 

\subsection{Emotional Support Conversation}

Emotional Support Conversation (ESC) is a task aimed at alleviating users' negative emotions (e.g., anxiety, depression), where \textbf{supporters} assist \textbf{seekers} in managing emotions triggered by issues like work crises or interpersonal conflicts. Typical ESC datasets include ESConv \cite{liu2021ESconv}, DailyDialog \cite{li-etal-2017-dailydialog} and EmpatheticDialogues \cite{rashkin-etal-2019-towards}. A typical ESC dataset usually has the following attributes:

\begin{itemize}
\item \textbf{Description}: a brief introduction before each conversation that provides insight into the current situation of the help-seeker, revealing their facing challenges.
\item \textbf{Query}: the seeker's query typically seeks emotional help. 
\item \textbf{Emotion}: can be emotion types or intensities, which capture the psychological state of the help-seeker.
\item \textbf{Strategy}: the support strategy of response selected for the current turn based on the seeker’s emotional state. 
\item \textbf{Response}: the supporter’s formal response, based on the above contexts.
\end{itemize}

In the practice, ESConv and DailyDialog\footnote{Strategy is instead named as `act' in DailyDialog.} have exactly the aforementioned architecture, while EmpatheticDialogues lacks the Strategy. Details can refer to Appendix \ref{appendix:dataset_detail}. Table \ref{tab:ex_ESConv} exhibits a detailed example of ESConv.

\begin{table*}[t!]
    \centering
    \small
    \begin{tabular}{c|l}
        \toprule
        \textit{Situation} & I hate my job but I am scared to quit and seek a new career. \\
        \midrule
        \textit{Query} & \makecell[l]{\textit{\{history\}} \\
        \makecell[l]{\textit{seeker:} Seriously! \\ What I am scare of now is how to secure another job.}}  \\
\midrule
        \textit{Emotion} & \textcolor{red}{Anxiety} (intensity: 5) \\
        \midrule
       \textit{Strategy} & \makecell[l]{\textcolor{blue}{Reflection of feelings}}   \\
\midrule
      \textit{Response} & \makecell[l]{\textit{supporter:} I can feel your pain just by chatting with you.}   \\
        \bottomrule
    \end{tabular}
    \caption{An example of ESConv.} 
    \label{tab:ex_ESConv}
\end{table*}

\begin{table*}[htbp!]
\centering
\small
\resizebox{0.99\textwidth}{!}{
\begin{tabular}{l p{10cm}p{2cm}}
      \toprule[1.5pt]
        \textbf{Stage} & \textbf{Prompt}& \textbf{Response} \\ 
        \midrule[1pt]
        \begin{tabular}[t]{@{}l@{}}
              Emotion \\ identification
        \end{tabular} & \begin{tabular}[t]{p{9.75cm}p{4.25cm}}
                    \begin{tabular}[t]{p{9.75cm}}
                $\langle$ History $\rangle$: $\{context\}$ \\
                Please propose the experience type of the history, the emotion, and the problem type of the "seeker" based on the history above.\\
            \end{tabular} & 
                \begin{tabular}[t]{p{4.25cm}}
                    $\langle$ Emotion  $\rangle$:\\ $\{emotion\}$ \\
                \end{tabular}
            \end{tabular} \\
        \midrule
        \begin{tabular}[t]{@{}l@{}}
              Strategy \\ determination
        \end{tabular} & \begin{tabular}[t]{p{9.75cm}p{4.25cm}}
                    \begin{tabular}[t]{p{9.75cm}}
                $ \langle $ History $ \rangle $: $\{context\}$ \\
                $ \langle $ Emotion $ \rangle $: $\{emotion\}$ \\
                Please choose the supporter's reply rule based on the above conversation history and state from the following options: \textit{Question}, \textit{Restatement or Paraphrasing}, \textit{Reflection of feelings}, \textit{Self-disclosure}, \textit{Affirmation and Reassurance}, \textit{Providing Suggestions}, \textit{Information}, \textit{Others}\\
            \end{tabular} & 
                \begin{tabular}[t]{p{4cm}}
                    $ \langle $ Strategy  $ \rangle $:\\ $\{strategy\}$ \\
                \end{tabular}
            \end{tabular} \\
        \midrule
        \begin{tabular}[t]{@{}l@{}}
              Response \\ generation
        \end{tabular} & \begin{tabular}[t]{p{10cm}p{4.25cm}}
                    \begin{tabular}[t]{p{9.75cm}}
                $ \langle $ History  $ \rangle $: $\{context\}$ \\
                $ \langle $ Emotion  $ \rangle $: $\{emotion\}$ \\
                $ \langle $ Strategy  $ \rangle $: $\{strategy\}$ \\
               Please respond appropriately as a supporter according to the rule.\\
            \end{tabular} & 
                \begin{tabular}[t]{p{4.25cm}}
                    $ \langle $ Response  $ \rangle $:\\ $\{response\}$ \\
                \end{tabular}
        \end{tabular} \\
    \bottomrule[1.5pt]
\end{tabular}
}
\caption{Prompts of {\ModelName} employed for action generations.}
\label{tab:prompt}
\end{table*}

\subsection{Task Definition}

The problem of emotional-support conversation (ESC) can be characterized by an interleaved sequence of seeker and supporter utterances, 
\begin{equation}
    ( u^k_t, u^p_t )_{t = 0, \dots, T}
\end{equation}
in which $u$ denotes the utterance, $T$ is the total number of conversation turn, and $k$, $p$ represent seeker and supporter, respectively.

To strengthen emotional-support performance, recent LM-based studies \cite{liu2021ESconv} enhance the data content by augmenting the set of support strategies $\mathcal{G}$ and seeker emotions $\mathcal{E}$. For each conversation session, the background description ($desc$) can also be annotated on the session level. Such an augmented ESC can then be described as
\begin{align}
    & desc, \{u^k_t, e_t, g_t, u^p_t \}_{t = 0, \dots, T} \label{eq:esc_seq}, \quad g \in \mathcal{G}, e \in \mathcal{E} 
\end{align}

At a specific time $t$, we denote the previous conversation history as
\begin{align}
    h_t = desc, \{ u^k_k, e_k, g_k, u^p_k \}_{k = 0, \dots, t-1} 
\end{align}
Then the ESC sample at time $t$ can be alternatively expressed as $\{ h_t, u^k_t, e_t, g_t, u^p_t \}$.

\begin{table*}[htbp!]
\centering
\small
\begin{tabular}{cl|ccccc| ccccc}
    \toprule
        & dataset $\rightarrow$ & \multicolumn{5}{c|}{Esconv} & \multicolumn{5}{c}{DailyDialog}\\ 
    \midrule
    & methods $\downarrow$ & $F1$ $\uparrow$ & $bias$ $\downarrow$ & B-2 $\uparrow$ & R-L $\uparrow$ & D-2 $\uparrow$  & $F1$ $\uparrow$ & $bias$ $\downarrow$  & B-2 $\uparrow$ & R-L $\uparrow$ & D-2 $\uparrow$ \\
    \midrule    
    \multicolumn{12}{l}{\textit{GPT4}} \\ 
    \multirow{4}{*}{\rotatebox{90}{\tiny Prompt\,}} 
    & \;+ Direct &  18.80 & 1.27 & 3.50 & 10.62 & 25.39 & 38.80 & 0.20 & 3.95 & 12.31 & 43.18 \\
    & \;+ Direct-Refine & 16.41 & 1.05 & 3.89 & 12.20 & 27.53 & 41.25 & 0.08 & 3.59 & 10.95 & 46.21 \\
    & \;+ Self-Refine & 20.20 & 1.03 & 5.58 & 11.80 & 24.29 & 29.64 & 0.95 & 3.61 & 11.12 & 45.97 \\
    & \;+ CoT & 15.12 & 1.08 & 3.51 & 10.59 & 21.88 & 33.92 & 0.60 & 3.89 & 11.88 & 44.35 \\
    \midrule
    \multicolumn{12}{l}{\textit{Llama3.1-8B-Instruct}} \\ 
    \multirow{4}{*}{\rotatebox{90}{\tiny Prompt\,}}
    & \;+ Direct & 10.26 & 1.61 & 3.47 & 10.64 & 33.45 & 18.03 & 1.66 & 3.35 & 10.33 & 44.74 \\ 
    & \;+ Direct-Refine & 11.07 & 1.27 & 3.10 & 6.13 & 14.22 & 28.28 & 0.70 & 2.56 & 8.70 & 43.67 \\
    & \;+ Self-Refine & 13.61 & 1.92 & 3.34 & 9.71 & 14.61 & 22.15 & 1.18 & 2.40 & 7.75 & 34.01 \\
    & \;+ CoT & 10.38 & 1.69 & 3.16 & 10.50 & 30.38 & 29.99 & \underline{0.27} & 1.78 & 6.00 & 45.26 \\
    \cline{2-12}
    \multirow{4}{*}{\rotatebox{90}{ \tiny Finetune\,}} 
    & \;+ SFT & \underline{22.70} & 0.84 & \underline{7.65} & \underline{17.30} & 39.18 & 44.82 & 0.82 & \bf 6.81 & \bf 18.52 & 43.36 \\
    & \;+ CoT + SFT  & 17.70 & 1.35 & 6.51 & 15.00 & 34.96 & \underline{44.90} & 0.66 & \underline{6.61} & \underline{18.07} & 42.87 \\
    & ArCher & 19.60 & \bf \textcolor{blue}{0.50} & 5.30 & 13.10 & \bf \textcolor{blue}{54.80} & 42.67 & \bf\textcolor{blue}{0.21} & 5.17 & 14.35 & \bf \textcolor{blue}{55.16} \\ 
    & \textbf{\ModelName} (ours) & \bf 23.30 & \underline{0.69} & \bf {8.52} & \bf \textcolor{blue}{19.20} & \underline{43.65} & \bf 46.02  & 0.55 & 4.58 & 17.32 & \underline{53.70} \\
    \midrule
    \multicolumn{12}{l}{\textit{Llama3.3-70B-Instruct}} \\
    \multirow{4}{*}{\rotatebox{90}{\tiny Prompt\,}}
    & \;+ Direct & 12.19 & 2.07 & 3.06 & 10.16 & 35.65 & 22.57 & 1.33 & 3.50 & 11.20 & 44.33 \\
    & \;+ Direct-Refine & 14.83 & 1.93 & 2.54 & 9.40 & 15.5 & 34.52 & 0.42 & 4.16 & 12.84 & 50.64 \\
    & \;+ Self-Refine & 16.15 & 1.76 & 2.97 & 10.12 & 14.84 & 29.63 & 0.95 & 3.40 & 10.83 & 33.62 \\
    & \;+ CoT & 7.83 & 2.35 & 1.85 & 9.67 & 30.62 & 33.92 & \bf 0.60 & 2.09 & 7.62 & 30.72 \\
    \cline{2-12}
    \multirow{3}{*}{\rotatebox{90}{ \tiny Finetune\,}} 
    & \;+ SFT & \underline{23.82} & \underline{0.90} & \underline{8.25} & \bf 18.50 & \underline{41.12} & \underline{47.41} & \underline{0.62} & \bf\textcolor{blue}{7.80} & \underline{20.13} & \underline{45.87} \\
    & \;+ CoT + SFT  & 17.76 & 0.98 & 6.25 & 14.95 & 36.6 & 46.74 & \textbf{0.60} & \underline{7.74} & 19.77 & 44.92 \\
    & \textbf{\ModelName} (ours) &  \bf\textcolor{blue}{25.72} &  \bf  0.81 &   \bf\textcolor{blue}{8.62} & \underline{17.60} &\bf 47.73 & \bf\textcolor{blue}{48.85} & 0.63 & 6.62 & \bf\textcolor{blue}{23.3} & \bf 47.98\\
    \bottomrule
\end{tabular}
\caption{ID results on ESConv and DailyDialog. For each column, values in \textbf{bold} indicate the best results, while values with \underline{underline} indicate the second-best results. The best results across all base sizes are marked in \textcolor{blue}{blue}.
}
\label{tab:methodology_results}
\end{table*}

\subsection{Framework}

Our {\ModelName} can be formally described as a 4-tuple $(\mathcal{S}, \mathcal{A}, \mathcal{C}, \mathcal{\delta})$, where $\mathcal{S} = \{ s_0, s_1, s_2, s_3 \}$ is the set of states in which 

    \begin{itemize}
    \item $s_0 = [h, u^k]$: the seeker utterance conditioned by the conversation history $h$
    \item $s_1 = [h, u^k, e]$: the user's emotional state has been recognized as $e$
    \item $s_2 = [h, u^k, e, g]$: the supporter has determined the current strategy $g$ based on available information
    \item $s_3 = [h, u^k, e, g, u^p]$: the supporter has generated the utterance $u^p$
    \end{itemize}
$\mathcal{A} = \{ u^k, e, g, u^p \}$ represents the set of actions within the state machine;
$\mathcal{C} = \{ desc \}$ is the set of job descriptions;  
$\mathcal{\delta}: \mathcal{S} \times \mathcal{A} \rightarrow \mathcal{S}$ is the transition function, defined as follows 
    \begin{itemize}
    \item $\mathcal{\delta}(s_{0,t}, e_t) = s_{1,t}$
    \item $\mathcal{\delta}(s_{1,t}, g_t )= s_{2,t}$
    \item $\mathcal{\delta}(s_{2,t}, u^p_t) = s_{3,t}$
    \item $\mathcal{\delta}(s_{3,t}) \xrightarrow[]{seeker} s_{0,t+1}$
    \end{itemize}
Within the last type of transition, the seeker provides the utterance $u^k$ (which is out of the control of {\ModelName}), then the next-step conversation history is stepwise appended by $h_{t+1} = h_t \cup [u^p_t, u^k_t]$.

\subsection{Implement on Large Language Models}

To determine the action given the current state, \textit{i.e.}, $\mathcal{S} \rightarrow \mathcal{A}$, we utilized a unified finetuned LLM which conducts different action generations:
\begin{itemize}
\setlength{\itemsep}{0em}
    \item $e = \text{LLM}(desc, s_0)$: classify the user emotion state 
    \item $g = \text{LLM}(desc, s_1)$: plan the optimal strategy based current situation
    \item $u^p = \text{LLM}(desc, s_2)$: generate the appropriate response guided by strategy
\end{itemize}
in which the LLM prompt is formulated by concatenating the contents of descriptions and all components of the corresponding states. Table \ref{tab:prompt} shows the detailed prompts. We denote this inference paradigm by 
\begin{equation}
    s_0 \Rrightarrow e \Rrightarrow g \Rrightarrow u^p \label{eq:inference}
\end{equation}
where each symbol $\Rrightarrow$ represents a single inference of LLM. We transform the original datasets into the multi-turn dialogue format. Finetuning is finally conducted on the mixture of datasets with equal fractions. 

\section{Experiment}

In this section, we first briefly introduce the implementation details, then the baselines and evaluation methods, and lastly detailed experimental results. Further details are provided in the Appendix.

\subsection{Setting}

Llama3.1-8B-Instruct and Llama3.3-70B-Instruct \cite{llama3modelcard} are employed as the backbones. Training is conducted on LlamaFactory \cite{zheng2024llamafactory}, with a learning rate of $1.0e-6$, window length of 1024, and epoch of 2. Batch size is set to 128. Experiments are conducted on 8 A100 80GB PCIe GPUs.

Since our methodology relies on fine-tuning of datasets with annotated strategies, we employ ESConv and DailyDialog as in-domain (ID) evaluation, and EmpatheticDialogues as out-of-domain (OOD) evaluation. For the ID test, both strategy- and response-related results can be provided. For the OOD test, we evaluated the zero-shot performance of the model fine-tuned on ESConv, with only response-related results provided.

\begin{table*}[htbp!]

\centering
\small
\begin{tabular}{c|ccccc|cccc}
    \toprule
    method $\rightarrow$ & \multicolumn{5}{c|}{prompting} & \multicolumn{4}{c}{finetuning} \\
    \cline{1-1} \cline{2-6} \cline{7-10}
    metric $\downarrow$ & Direct & D-Refine & S-Refine & CoT & \textbf{\ModelName} & SFT & C+SFT & ArCHer & \textbf{\ModelName} \\
    \midrule
    B-2 & 3.09 & 2.56 & 3.08 & 2.91 & \bf 3.33 & 3.74 & 3.75& 3.84 & \bf \textcolor{blue}{4.98} \\ 
    R-L & 9.91  & 9.12 & 9.91 & 9.79 & \bf 10.80 & 14.08 & 14.06 & 10.75 & \bf \textcolor{blue}{14.59} \\ 
    D-2 & 25.23 & 22.32 & 25.20 & 32.65 & \bf 33.37 & 25.73 & 27.30 & \bf \textcolor{blue}{45.19} & 40.39 \\
    \bottomrule
\end{tabular}
\caption{OOD results of automatic metrics on EmpatheticDialogues. D-Refine, S-Refine, and C+SFT denote Direct-Refine, Self-Refine, and CoT+SFT, respectively.
}
\label{tab:auto_empatheticdialogues}
\end{table*}

\begin{table*}[htbp!]
\centering
\small
\begin{tabular}{l|ccccccc}
    \toprule
    \multicolumn{1}{c|}{\multirow{2}[4]{*}{Method}} & \multicolumn{7}{c}{Human Annotation} \\
\cmidrule{2-8}    
 & Fluency  & Emotion & Acceptance & Effectiveness & Sensitivity & Alignment &  Satisfaction \\ 
    \toprule   
    original dataset &  3.46 (1.27) & 3.61 (1.17) & \underline{3.42} (1.19) & 3.14 (1.19) & 3.51 (1.28) & 3.17 (1.30) & 3.33 (1.19) \\
    \midrule
    \multicolumn{8}{l}{\textit{model on larger sizes}} \\
    Llama3.3-70B-inst  & 3.05 (1.36) & 3.33 (1.20) & 2.76 (1.17) & 3.08 (1.36) & 3.22 (1.09) & 3.11 (1.48) & 3.09 (1.14) \\
    GPT-4  & 3.53 (1.31) & 3.71 (1.09) & 3.45 (1.36) & 3.28 (1.10) & 3.48 (1.38) & \underline{3.52} (1.24) & 3.40 (1.22) \\
    GPT-4o  & 3.62 (0.98) & \textbf{3.84} (1.21) & \textbf{3.71} (1.15) & \underline{3.59} (1.30) & \underline{3.64} (1.14) & 3.21 (1.10) & \underline{3.55} (1.12) \\
    \midrule
    \multicolumn{8}{l}{\textit{model on 8B}} \\
    Direct  &  2.95 (1.43) & 3.00 (1.30) & 2.60 (1.33) & 2.40 (1.01) & 2.70 (1.17) & 2.70 (1.33) & 2.60 (1.30) \\
    \;+ Direct-Refine & 3.09 (1.14) & 3.09 (1.35) & 2.73 (1.26) & 2.91 (1.23) & 2.91 (1.29) & 2.82 (0.97) & 2.84 (1.32) \\
    \;+ Self-Refine & 3.10 (1.30) & 3.15 (1.2) & 2.80 (1.44) & 2.72 (1.17) & 2.90 (1.13) & 2.81 (1.36) & 2.80 (1.12) \\
    \;+ CoT  & 3.08 (1.20) & 3.08 (1.34) & 2.83 (1.16) & 2.67 (1.05) & 3.00 (1.14) & 2.83 (1.16) & 2.83 (1.30) \\
    \;+ {\vanillaFT} &  3.15 (1.19) & 3.40 (1.37) & 2.70 (1.10) & 2.73 (1.37) & 2.91 (1.33) & 3.32 (1.29) & 2.90 (1.30) \\
    \;+ CoT+SFT  &  \underline{3.67} (1.20) & 3.61 (1.16) & 3.22 (1.25) & 2.67 (1.31) & 3.56 (1.26) & 3.35 (1.26) & 3.45 (1.20) \\
    \textbf{\ModelName} (ours) &  \textbf{3.80} (1.16) & \underline{3.55} (1.28) & 3.40 (1.39) & \textbf{3.70} (1.21) & \textbf{3.80} (1.15) & \textbf{3.70} (1.09) & \textbf{3.65} (1.12) \\
    \bottomrule
    \end{tabular}%
\caption{Averaged Human scores of response quality on ESConv, DailyDialog, and EmpatheticDialogues. Values with bold indicate the best results, while values with underline indicate the second-best results. Standard deviations are in parentheses.}
\label{tab:response_quaility}
\end{table*}

\subsection{Baselines}

Besides \textbf{Direct} inference of the base model, we consider the following baselines:


\noindent (1) \textit{Prompt-based}: we test \textbf{Direct-Refine}, in which LLM first generates the response, then revises it immediately; \textbf{Self-Refine} \cite{Madaan2023SelfRefine}, which refines the response with the response of the second time inference; and \textbf{CoT} \cite{wei2022chain,li2024enhancing}, in which we encourage LLM to sequentially generate \textit{emotion}, \textit{strategy}, and finally the response.



\noindent (2) \textit{Finetuning-based}: besides the conventional \textbf{{\vanillaFT}}, we also test \textbf{CoT + SFT}, in which we first use CoT to infer the chained responses, then finetune on this sampled dataset; and finally, we test a reinforcement learning (RL) baseline called \textbf{ArCher} \cite{10.5555/3692070.3694644}, which plans the strategy and generate the response by a hierarchical policy.



\subsection{Evaluation Methods}
\label{sec:evaluation}
\paragraph{Automatic metrics.} We use the famous Macro-F1 ($F1$) to indicate classification and decision accuracies. Similar to \citet{kang-etal-2024-large}, we also investigate the preference $bias$ based on the Bradley-Terry model~\cite{bradley1952btmodel}, with smaller $bias$ denoting more balanced strategy determination. 

For response quality, we employ the famous BLEU-2 (\textbf{B-2}), Rouge-L (\textbf{R-L}) and Dist-2 (\textbf{D-2}). Theoretically, B-2 and R-L represent the similarity of the model response with respect to the ground-truth response, while D-2 represents the diversity of the model response itself. 


\paragraph{Human evaluation.} Responses are rated across dimensions of \textit{Acceptance}, \textit{Effectiveness}, \textit{Sensitivity}, \textit{Fluency}, and \textit{Emotion}, and the ultimate purpose, seeker's \textit{Satisfaction}. Detailed annotation principles can refer to \citet{kang-etal-2024-large}. Scores are averaged over ESConv, DailyDialog, and EmpatheticDialogues.




\paragraph{LLM-as-a-Judge} We used GPT-4o to compare different responses and provide win-tie-lose suggestions, similar to \citet{madani2024steeringconversationallargelanguage}. One can refer to Table \ref{tab:prompt_comparison_score} for the detailed evaluation prompt.

\subsection{Main Results}

\paragraph{ID results.} Table \ref{tab:methodology_results} lists results on Esconv and DailyDialog, including classification metrics such as $F1$ and $bias$, and generative metrics including B-2, R-L, and D-2. {\ModelName} significantly outperforms prompting and finetuning methods in both strategy determination and response quality, on both sizes of 8B and 70B. {\ModelName} also surpasses the prompting baselines on GPT4, which has a much larger model size. Finally, {\ModelName} achieves a similar performance to ArCher, which is computationally expensive. In Appendix \ref{sec:significance_results}, we further exhibit the statistical significance of our results in Table \ref{tab:significance_results_auto}.







\begin{table*}[htbp!]
\centering
\small

\resizebox{0.99\textwidth}{!}{
\begin{tabular}{l p{3cm}p{12.5cm}}
      \toprule[1.5pt]
        \textbf{Characters} & \textbf{Strategy}& \textbf{Response} \\ 
        \midrule
        \textcolor{brown}{supporter}& - &\textcolor{brown}{ Hi! Hope you are doing well?} \\
        \midrule
        \textcolor{brown}{seeker}& - & \textcolor{brown}{Hi, yes am doing well }\\
        \midrule
        \textcolor{brown}{supporter}& - & \textcolor{brown}{How can I help you today?}\\
        \midrule
        \textcolor{brown}{seeker}& - & \textcolor{brown}{I had a drinking challenge which has been affecting my marriage for quit sometime now }\\
        \midrule
        supporter& \makecell[l]{Restatement or\\ Paraphrasing.} & So as far as I understand that you had issues with your wife due to your drinking. Am I right?\\
        Direct & \makecell[l]{Restatement or\\ Paraphrasing.} & So, it sounds like there's been a bit of tension in your marriage lately due to your drinking habits, and it even led to a confrontation that's caused a rift between you and your wife. Can you tell me more about what's been going on with your drinking? \\
        \textbf{\ModelName} & \textbf{\makecell[l]{Restatement or\\ Paraphrasing.}} & \textbf{I see. So you are having issues with your spouse due to your drinking habit. Is it correct?}\\
        \midrule
        \textcolor{brown}{seeker}& - & \textcolor{brown}{Yes that is true}\\
        \midrule
        supporter& \makecell[l]{Affirmation and\\ Reassurance} & That is really a serious problem. I know it must be very tough time for you. Hope you are trying your best to save your marriage.\\
        Direct & \makecell[l]{Restatement or\\ Paraphrasing.} & Just to make sure I understand correctly, it sounds like there was a confrontation with your wife about your drinking habits, and now you two aren't speaking. Is that a fair summary of what happened? \\
        \textbf{\ModelName} &\textbf{\makecell[l]{Affirmation and\\ Reassurance}} &\textbf{ I can understand how difficult it must be for you. But don't worry, I am here to help you out.}\\
        \midrule
        \textcolor{brown}{seeker}& - & \textcolor{brown}{Yes I have really reduced my drinking to ensure I save my marriage} \\
        \midrule
        supporter& \makecell[l]{Reflection  of\\ Feelings} & That's is a great initiative. I know it is tough for you but not impossible. I can feel how hard you are trying t save your relationship. I appreciate you for that.\\
        Direct & \makecell[l]{Restatement or\\ Paraphrasing.} & Just to make sure I understand correctly, you've been trying to cut down on your drinking to address your wife's concerns, but despite your efforts, the issue still seems to be causing tension in your marriage, and now you're not talking to each other. Is that a correct summary? \\
        \textbf{\ModelName} & \textbf{\makecell[l]{Affirmation and\\ Reassurance}} & \textbf{I can feel how much you are struggling with this issue. I think you should try to find some professional help.}\\
        \bottomrule
\end{tabular}
}
\caption{Typical cases generated by {\ModelName} in ESConv. Contents of the original dataset are brown; contents of the {\ModelName} are bolded.}
\label{tab:case}
\end{table*}

\paragraph{OOD performance.} Table \ref{tab:auto_empatheticdialogues} further shows the OOD results on EmpatheticDialogues. {\ModelName} still outperforms all baselines, except ArCHer on D-2, which may be because ArCHer's trial-and-error methodology may benefit the response diversity. This positive observation on the OOD study further validates the generalization capability of {\ModelName} in unfamiliar scenarios and contexts.

\paragraph{Human evaluations.} In Table \ref{tab:response_quaility}, {\ModelName} still surpasses all other baselines, even models on much larger sizes (Llama3.3-70B-Instruct, GPT4, GPT4o), demonstrating stronger emotional support capabilities. Table \ref{tab:significance_results_human} in Appendix \ref{sec:significance_results} further provides the corresponding p-values.





\paragraph{Typical cases.} Table \ref{tab:case} provides a typical example of {\ModelName}. It can be observed that {\ModelName} can predict more accurate strategies compared to the base model, and finally generate high-quality responses.

\begin{table*}[t!]
\centering
\small
\begin{tabular}{l | cccc | ccc}
\toprule
     experiment & \multicolumn{4}{c|}{Comp. to ground-truth} & \multicolumn{3}{c}{vs SFT} \\
     \cline{1-1} \cline{2-5} \cline{6-8}
     method    & $F1 \uparrow$  &  $bias \downarrow$   &  B-2 $\uparrow$  &  R-L $\uparrow$  & win $\uparrow$ & tie & lose $\downarrow$ \\ 
\midrule
 w/o $e$  & 22.32 & 0.70 & 8.23 & 19.04 \\ 
 w/o $g$  & N\slash A & N\slash A & 6.69 & 16.11 \\ 
 w/o $desc$ & 18.36 & 1.32 & 7.57 & 17.42 \\ 
 w/o $FT$ & 11.15 & 0.81 & 4.11 & 11.85\\  
 w/ multi-model & 22.30 & 0.49 & 5.71 & 15.40 & 51.8\%  & 6.3\%  & 41.8\% \\
 w/ single-turn & 23.70 & \bf 0.41 & 5.88 & 15.30  & 53.5\%  & 7.4\%  & 39.2\% \\
\bf {\ModelName} & \bf 23.33 & 0.69 & \bf 8.52 & \bf 19.22 & \bf 59.5\%  & 11.1\%  & \bf 29.4\%  \\ 
\bottomrule
\end{tabular}
\caption{Ablation study on the basis of Llama-3-8B-Instruct. Method w/o $g$ does not have strategy-related results since it excludes reasoning of strategies.}
\label{tab:ablation_results}
\end{table*}


\paragraph{Ablation study.} We conducted the following ablation studies on ESConv:

\begin{itemize}
    \setlength{\itemsep}{0em}
    \item w/o $e$: do not consider the seeker's emotion $e$.
    \item w/o $g$: do not consider the strategy $e$.
    \item w/o $desc$: do not consider the session description.
    \item w/o FT: instead of finetuning, evaluate the model in the zero-shot way.
    \item w/ multi-model: train separate models for each state-action transition.
    \item w/ single-turn: train the model on the single-turn samples.
\end{itemize}

In Table \ref{tab:ablation_results}, we first calculate the automatic metrics of these variants, including $F1$, $bias$, B-2, and R-L, all of which are computed based on the ground-truth strategies and responses in the original datasets. While w/ single-turn has a lower $bias$ than {\ModelName}, generally {\ModelName} still has the best performance, indicating each key component of {\ModelName} is effective. 

In addition, we also provide the win-tie-lose rates of w/ multi-model, w/ single-turn, and the formal {\ModelName} versus {\vanillaFT}, all of which are pairwise evaluated by GPT-4o. Both w/ multi-model and w/ single-turn still have higher winning rates than {\vanillaFT}, but still worse than {\ModelName}.


\subsection{Further Discussions}

\begin{figure*}[htbp!]
\centering
    \includegraphics[width=\linewidth]{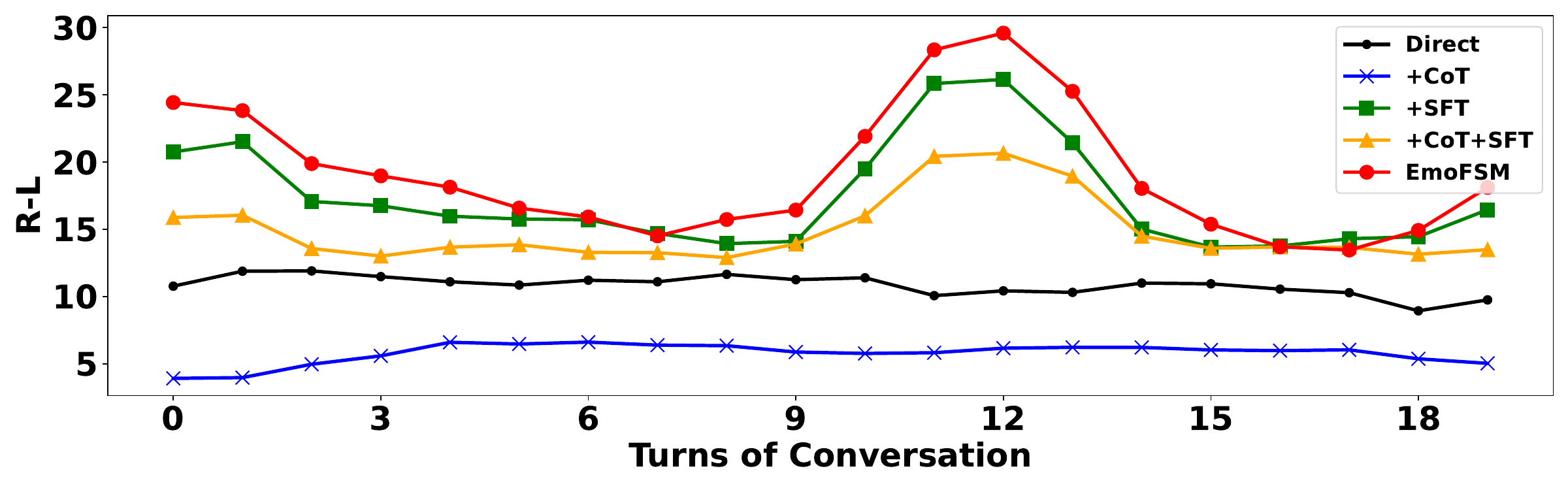}
    \caption{R-L changes on conversation turns. Results are collected from ESConv.}
    \label{fig:performance_turns}
\end{figure*}

\paragraph{Evolution on conversation turns.} Fig. \ref{fig:performance_turns} shows the curves of R-L as the number of conversation turns increases. It can be observed that the performance of {\ModelName} does not degrade as the conversation continues and may even be stronger when the number of turns is close to 12. Although finetuning-based baselines exhibit similar trends, they are generally worse than {\ModelName}, indicating the effectiveness of FSM for long-horizon planning.

\begin{figure}[htpb!]
    \centering
    \includegraphics[width=\linewidth]{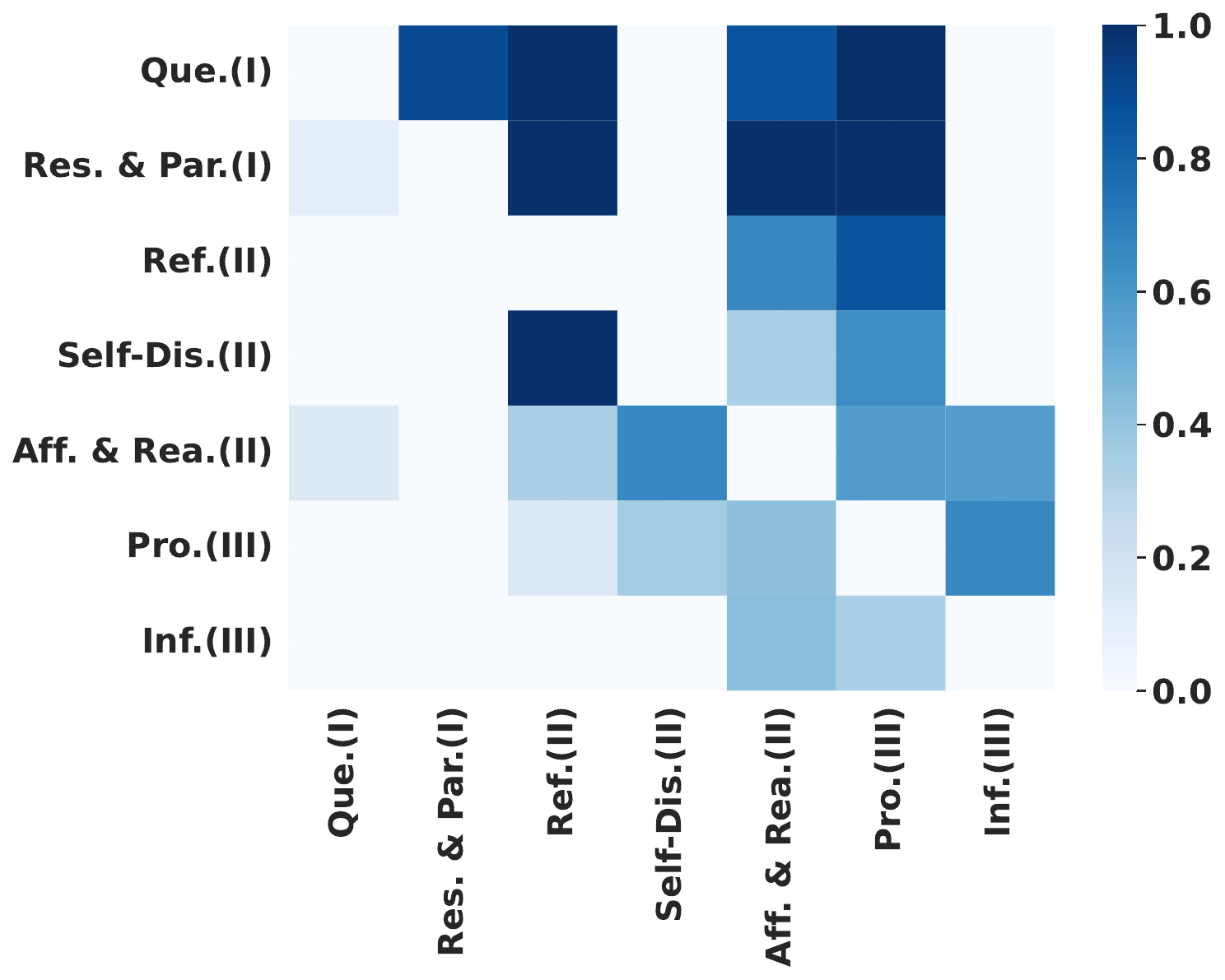}
    \caption{The transition distribution between different strategies. Each grid represents the transition from the row-strategy to the column-strategy. Strategies are marked by their ESC stages (marked by I, II, and III) and ordered correspondingly.}
    \label{fig:strategy_trainsition}
\end{figure}


\paragraph{Analysis on Strategies} Fig. \ref{fig:strategy_trainsition} further exhibits the strategy transitions within consecutive turns in the same conversation, determined by {\ModelName}. For each grid, the color represents the fraction of the transitions from the row-strategy to the column-strategy. According to \cite{liu2021ESconv}, the strategies can be related to three supporting stages (Exploration, Comforting, and Action). Therefore, we mark each strategy with the corresponding stage I, II, and III and order the strategies accordingly. As a result, a more reasonable strategy transition may happen from earlier to later stages \textit{i.e.}, the upper-triangle part of the matrix. Obviously, such a pattern can be observed from Fig. \ref{fig:strategy_trainsition}, indicating that the strategy determination of {\ModelName} is aligned with the `Exploration, Comforting and Action' paradigm.

\begin{table*}[htbp!]

\centering
\caption{Comparison evaluations by GPT-4o between different {\ModelName} variants. $\Rrightarrow$ represents the LLM inference. Results in the upper triangle represent the (win, tie, lose) rates of row variant VS column variant.}
\resizebox{0.98\textwidth}{!}{

\begin{tabular}{ccccc}
\hline
(win, tie, loss) Rate   & $s_0 \Rrightarrow e \Rrightarrow g \Rrightarrow u^p$ & $s_0, e \Rrightarrow g \Rrightarrow u^p$ & $s_0, e, g \Rrightarrow u^p$ & $s_0, e, g, u^p$ \\
\hline
$s_0 \Rrightarrow e \Rrightarrow g \Rrightarrow u^p$   & - & (47\%, 8\%, 45\%) &  (53\%, 7\%, 40\%) & (54\%, 4\%, 42\%) \\
$s_0, e \Rrightarrow g \Rrightarrow u^p$   & -   & -  & (51\%, 9\%, 40\%) & (54\%, 5\%, 41\%) \\
$s_0, e, g \Rrightarrow u^p$ & -   & - & -  & (46\%, 18\% ,36\%) \\
$s_0, e, g, u^p$  & -   & - & - & - \\
\hline
\end{tabular}}
\label{tab:variant_ablation}
\end{table*}

\paragraph{Effectiveness of self-reasoning.} {\ModelName} has three inference steps, \textit{i.e.}, $s_0 \Rrightarrow e \Rrightarrow g \Rrightarrow u^p$  as specified in Equation (\ref{eq:inference}). One may argue whether all these inference steps are necessary, and how variants with fewer inference steps perform. To validate this consideration, we test the following variants by assuming some of the state components could be obtained directly from the original dataset, while LLM generates the rest:

\begin{itemize}
    \item $s_0, e \Rrightarrow g \Rrightarrow u^p$: first generate $g$ based on $s_0$ and $e$, then generate the utterance $u^p$.
    \item $s_0, e, g \Rrightarrow u^p$: generate the utterance $u^p$ once grounded by $s_0, e, g$.
    \item $s_0, e, g, u^p$: use the original annotations from the raw dataset.
\end{itemize}

Table \ref{tab:variant_ablation} shows the comparison evaluation results between different variants. It shows that the formal {\ModelName} ($s_0, e \Rrightarrow g \Rrightarrow u^p$) is better than all other variants; and the more LLM inferences, the better the performance. This phenomenon indicates that our FSM mechanism boosts the LLM to reasonably bootstrap the conversation situation and make remarkable decisions. Although FSM methods need to fine-tune the LLM on the raw dataset, they outperform original dataset responses ($s_0, e, g, u^p$).

\section{Related Work}

\subsection{Emotional Support Conversations}

There have been substantial studies of LLMs on ESC. For example, \citet{liu2021ESconv} introduces ESConv and emphasizes strategy planning via the Helping Skills theory. \citet{cheng-etal-2022-improving} propose MultiESC, addressing multi-turn challenges using lookahead heuristics. \citet{zhao-etal-2023-transesc} models turn-level state transitions to improve support coherence, while \citet{zheng-etal-2023-augesc} augments ESC data using LLMs. \citet{zhang-etal-2023-ask} proposes Ask-Expert to enhance emotional dialogue with expert advice, and \citet{zheng-etal-2024-self} introduces a teacher-student framework (DRI) to improve diversity. \citet{li2024enhancing} combines CoT and emotion modeling in ECoT. \citet{kang-etal-2024-large} identifies LLM limitations and explores mitigation through external assistance.

In this work, we instead apply the finite state machine (FSM) to ESC tasks. Our methodology leverages the LLM's capability for situation recognition and support strategy determination, and subsequent response generation.

\subsection{Finite State Machine on NLP tasks}

There are previous attempts of FSM on Multi-hop Question Answering \cite{wang2024fsmfinitestatemachine,wang2024sgfsm} and reasoning \cite{liu2024SMoT}. \citet{wang2024fsmfinitestatemachine} applies FSM to Multi-hop Question Answering (MHQA) by iteratively decomposing complex questions into sub-questions and self-correcting errors during the process. SG-FSM \cite{wang2024sgfsm} further enhances this by dynamically deciding the next step in the reasoning process. It improves the model's ability to handle multi-hop reasoning tasks and reduce hallucination. SMoT \cite{liu2024SMoT} applies a framework similar to FSM to reasoning tasks. It addresses complex scenarios with multi-turn sub-questions and feedback loops to ensure more accurate results.

In contrast, our {\ModelName} applies FSM to Emotional Support Conversations (ESC). We use FSM to structure the model's reasoning process, allowing the FSM to adaptively adjust strategies to provide effective emotional responses, ensuring a smooth and empathetic dialogue. By enabling LLMs to infer the seeker's emotional state from the history of the conversation, we determine the specific strategy for responding, ultimately combining these strategies to generate the appropriate reply.



\section{Conclusion}

In this work, we propose the {\ModelName} framework, which leverages a Finite State Machine (FSM) to enhance Emotional Support Conversations (ESC).  {\ModelName} combines the dynamics of the seeker's emotions, the supporter's strategies, and the utterances of both sides, providing a more structured and context-aware approach to dialogue generation. Results confirm that our {\ModelName} framework outperforms prompt-based, supervised learning, or reinforcement learning baselines in ESC. In-depth analysis also indicates that {\ModelName} can perform long-horizon planning on conversation strategies, conducting reasonable strategy transitions consistent with the emotional support theories.

\clearpage
\newpage

\bibliography{custom}

\begin{thebibliography}{25}
\providecommand{\natexlab}[1]{#1}

\bibitem[{AI@Meta(2024)}]{llama3modelcard}
AI@Meta. 2024.
\newblock \href {https://github.com/meta-llama/llama3/blob/main/MODEL_CARD.md} {Llama 3 model card}.

\bibitem[{Bradley and Terry(1952)}]{bradley1952btmodel}
Ralph~Allan Bradley and Milton~E Terry. 1952.
\newblock Rank analysis of incomplete block designs: I. the method of paired comparisons.
\newblock \emph{Biometrika}, 39(3/4):324--345.

\bibitem[{Cheng et~al.(2022)Cheng, Liu, Li, Wang, Zhao, Liu, Liang, and Zheng}]{cheng-etal-2022-improving}
Yi~Cheng, Wenge Liu, Wenjie Li, Jiashuo Wang, Ruihui Zhao, Bang Liu, Xiaodan Liang, and Yefeng Zheng. 2022.
\newblock \href {https://doi.org/10.18653/v1/2022.emnlp-main.195} {Improving multi-turn emotional support dialogue generation with lookahead strategy planning}.
\newblock In \emph{Proceedings of the 2022 Conference on Empirical Methods in Natural Language Processing}, pages 3014--3026, Abu Dhabi, United Arab Emirates. Association for Computational Linguistics.

\bibitem[{Heaney and Israel(2008)}]{heaney2008social}
Catherine~A Heaney and Barbara~A Israel. 2008.
\newblock Social networks and social support.
\newblock \emph{Health behavior and health education: Theory, research, and practice}, 4(1):189--210.

\bibitem[{Kang et~al.(2024)Kang, Kim, Kwon, Moon, Cho, Yu, Lee, and Yeo}]{kang-etal-2024-large}
Dongjin Kang, Sunghwan Kim, Taeyoon Kwon, Seungjun Moon, Hyunsouk Cho, Youngjae Yu, Dongha Lee, and Jinyoung Yeo. 2024.
\newblock \href {https://doi.org/10.18653/v1/2024.acl-long.813} {Can large language models be good emotional supporter? mitigating preference bias on emotional support conversation}.
\newblock In \emph{Proceedings of the 62nd Annual Meeting of the Association for Computational Linguistics (Volume 1: Long Papers)}, pages 15232--15261, Bangkok, Thailand. Association for Computational Linguistics.

\bibitem[{Langford et~al.(1997)Langford, Bowsher, Maloney, and Lillis}]{langford1997social}
Catherine Penny~Hinson Langford, Juanita Bowsher, Joseph~P Maloney, and Patricia~P Lillis. 1997.
\newblock Social support: a conceptual analysis.
\newblock \emph{Journal of advanced nursing}, 25(1):95--100.

\bibitem[{Li et~al.(2017)Li, Su, Shen, Li, Cao, and Niu}]{li-etal-2017-dailydialog}
Yanran Li, Hui Su, Xiaoyu Shen, Wenjie Li, Ziqiang Cao, and Shuzi Niu. 2017.
\newblock \href {https://aclanthology.org/I17-1099/} {{D}aily{D}ialog: A manually labelled multi-turn dialogue dataset}.
\newblock In \emph{Proceedings of the Eighth International Joint Conference on Natural Language Processing (Volume 1: Long Papers)}, pages 986--995, Taipei, Taiwan. Asian Federation of Natural Language Processing.

\bibitem[{Li et~al.(2024)Li, Chen, Shao, Xie, Jiang, and Nie}]{li2024enhancing}
Zaijing Li, Gongwei Chen, Rui Shao, Yuquan Xie, Dongmei Jiang, and Liqiang Nie. 2024.
\newblock Enhancing emotional generation capability of large language models via emotional chain-of-thought.
\newblock \emph{arXiv preprint arXiv:2401.06836}.

\bibitem[{Lin(2004)}]{lin2004rouge}
Chin-Yew Lin. 2004.
\newblock Rouge: A package for automatic evaluation of summaries.
\newblock In \emph{Text summarization branches out}, pages 74--81.

\bibitem[{Liu et~al.(2024)Liu, Shuai, and Li}]{liu2024SMoT}
Jia Liu, Jie Shuai, and Xiyao Li. 2024.
\newblock \href {https://arxiv.org/abs/2312.17445} {State machine of thoughts: Leveraging past reasoning trajectories for enhancing problem solving}.
\newblock \emph{Preprint}, arXiv:2312.17445.

\bibitem[{Liu et~al.(2021)Liu, Zheng, Demasi, Sabour, Li, Yu, Jiang, and Huang}]{liu2021ESconv}
Siyang Liu, Chujie Zheng, Orianna Demasi, Sahand Sabour, Yu~Li, Zhou Yu, Yong Jiang, and Minlie Huang. 2021.
\newblock \href {https://doi.org/10.18653/v1/2021.acl-long.269} {Towards emotional support dialog systems}.
\newblock In \emph{Proceedings of the 59th Annual Meeting of the Association for Computational Linguistics and the 11th International Joint Conference on Natural Language Processing (Volume 1: Long Papers)}, pages 3469--3483, Online. Association for Computational Linguistics.

\bibitem[{Madaan et~al.(2023)Madaan, Tandon, Gupta, Hallinan, Gao, Wiegreffe, Alon, Dziri, Prabhumoye, Yang, Welleck, Majumder, Gupta, Yazdanbakhsh, and Clark}]{Madaan2023SelfRefine}
Aman Madaan, Niket Tandon, Prakhar Gupta, Skyler Hallinan, Luyu Gao, Sarah Wiegreffe, Uri Alon, Nouha Dziri, Shrimai Prabhumoye, Yiming Yang, Sean Welleck, Bodhisattwa~Prasad Majumder, Shashank Gupta, Amir Yazdanbakhsh, and Peter Clark. 2023.
\newblock Self-refine: Iterative refinement with self-feedback.
\newblock \emph{ArXiv}, abs/2303.17651.

\bibitem[{Madani et~al.(2024)Madani, Saha, and Srihari}]{madani2024steeringconversationallargelanguage}
Navid Madani, Sougata Saha, and Rohini Srihari. 2024.
\newblock \href {https://arxiv.org/abs/2402.10453} {Steering conversational large language models for long emotional support conversations}.
\newblock \emph{Preprint}, arXiv:2402.10453.

\bibitem[{Papineni et~al.(2002)Papineni, Roukos, Ward, and Zhu}]{papineni2002bleu}
Kishore Papineni, Salim Roukos, Todd Ward, and Wei-Jing Zhu. 2002.
\newblock Bleu: a method for automatic evaluation of machine translation.
\newblock In \emph{Proceedings of the 40th annual meeting of the Association for Computational Linguistics}, pages 311--318.

\bibitem[{Rains et~al.(2020)Rains, Pavlich, Lutovsky, Tsetsi, and Ashtaputre}]{rains2020support}
Stephen~A Rains, Corey~A Pavlich, Bethany Lutovsky, Eric Tsetsi, and Anjali Ashtaputre. 2020.
\newblock Support seeker expectations, support message quality, and supportive interaction processes and outcomes: The case of the comforting computer program revisited.
\newblock \emph{Journal of Social and Personal Relationships}, 37(2):647--666.

\bibitem[{Rashkin et~al.(2019)Rashkin, Smith, Li, and Boureau}]{rashkin-etal-2019-towards}
Hannah Rashkin, Eric~Michael Smith, Margaret Li, and Y-Lan Boureau. 2019.
\newblock Towards empathetic open-domain conversation models: A new benchmark and dataset.
\newblock In \emph{Proceedings of the 57th Annual Meeting of the Association for Computational Linguistics}, pages 5370--5381, Florence, Italy. Association for Computational Linguistics.

\bibitem[{Wang et~al.(2024{\natexlab{a}})Wang, He, Chen, Yang, Wang, Meng, Pan, and Sui}]{wang2024sgfsm}
Xiaochen Wang, Junqing He, Liang Chen, Reza Haf~Zhe Yang, Yiru Wang, Xiangdi Meng, Kunhao Pan, and Zhifang Sui. 2024{\natexlab{a}}.
\newblock \href {https://arxiv.org/abs/2410.17021} {Sg-fsm: A self-guiding zero-shot prompting paradigm for multi-hop question answering based on finite state machine}.
\newblock \emph{Preprint}, arXiv:2410.17021.

\bibitem[{Wang et~al.(2024{\natexlab{b}})Wang, He, yang, Wang, Meng, Pan, and Sui}]{wang2024fsmfinitestatemachine}
Xiaochen Wang, Junqing He, Zhe yang, Yiru Wang, Xiangdi Meng, Kunhao Pan, and Zhifang Sui. 2024{\natexlab{b}}.
\newblock \href {https://arxiv.org/abs/2407.02964} {Fsm: A finite state machine based zero-shot prompting paradigm for multi-hop question answering}.
\newblock \emph{Preprint}, arXiv:2407.02964.

\bibitem[{Wei et~al.(2022)Wei, Wang, Schuurmans, Bosma, Xia, Chi, Le, Zhou et~al.}]{wei2022chain}
Jason Wei, Xuezhi Wang, Dale Schuurmans, Maarten Bosma, Fei Xia, Ed~Chi, Quoc~V Le, Denny Zhou, et~al. 2022.
\newblock Chain-of-thought prompting elicits reasoning in large language models.
\newblock \emph{Advances in Neural Information Processing Systems}, 35:24824--24837.

\bibitem[{Zhang et~al.(2023)Zhang, Naradowsky, and Miyao}]{zhang-etal-2023-ask}
Qiang Zhang, Jason Naradowsky, and Yusuke Miyao. 2023.
\newblock Ask an expert: Leveraging language models to improve strategic reasoning in goal-oriented dialogue models.
\newblock In \emph{Findings of the Association for Computational Linguistics: ACL 2023}, pages 6665--6694, Toronto, Canada. Association for Computational Linguistics.

\bibitem[{Zhao et~al.(2023)Zhao, Zhao, Wang, and Qin}]{zhao-etal-2023-transesc}
Weixiang Zhao, Yanyan Zhao, Shilong Wang, and Bing Qin. 2023.
\newblock {T}rans{ESC}: Smoothing emotional support conversation via turn-level state transition.
\newblock In \emph{Findings of the Association for Computational Linguistics: ACL 2023}, pages 6725--6739, Toronto, Canada. Association for Computational Linguistics.

\bibitem[{Zheng et~al.(2023)Zheng, Sabour, Wen, Zhang, and Huang}]{zheng-etal-2023-augesc}
Chujie Zheng, Sahand Sabour, Jiaxin Wen, Zheng Zhang, and Minlie Huang. 2023.
\newblock {A}ug{ESC}: Dialogue augmentation with large language models for emotional support conversation.
\newblock In \emph{Findings of the Association for Computational Linguistics: ACL 2023}, pages 1552--1568, Toronto, Canada. Association for Computational Linguistics.

\bibitem[{Zheng et~al.(2024{\natexlab{a}})Zheng, Zhang, Zhang, Ye, Luo, and Ma}]{zheng2024llamafactory}
Yaowei Zheng, Richong Zhang, Junhao Zhang, Yanhan Ye, Zheyan Luo, and Yongqiang Ma. 2024{\natexlab{a}}.
\newblock \href {http://arxiv.org/abs/2403.13372} {Llamafactory: Unified efficient fine-tuning of 100+ language models}.
\newblock \emph{arXiv preprint arXiv:2403.13372}.

\bibitem[{Zheng et~al.(2024{\natexlab{b}})Zheng, Liao, Deng, Qin, and Nie}]{zheng-etal-2024-self}
Zhonghua Zheng, Lizi Liao, Yang Deng, Libo Qin, and Liqiang Nie. 2024{\natexlab{b}}.
\newblock Self-chats from large language models make small emotional support chatbot better.
\newblock In \emph{Proceedings of the 62nd Annual Meeting of the Association for Computational Linguistics (Volume 1: Long Papers)}, pages 11325--11345, Bangkok, Thailand. Association for Computational Linguistics.

\bibitem[{Zhou and Zanette(2024)}]{10.5555/3692070.3694644}
Yifei Zhou and Andrea Zanette. 2024.
\newblock Archer: training language model agents via hierarchical multi-turn rl.
\newblock In \emph{Proceedings of the 41st International Conference on Machine Learning}, ICML'24. JMLR.org.

\end{thebibliography}

\newpage

\appendix

\begin{table*}[htbp!] 

\centering
\small
\begin{tabular}{llccc}
\toprule
\multicolumn{2}{c}{Category} & ESConv (ID) & DailyDialog (ID) & EmpatheticDialogues (OOD)   \\
\midrule
\multicolumn{2}{l}{\# Sessions} & 1.3K & 13.1k& 2.5K \\
\multicolumn{2}{l}{\# Utterances} & 38K & 103.0k& 11.0K\\
\multicolumn{2}{l}{Average \# Utterances} & 28.9  & 7.9& 4.3\\
\multicolumn{2}{l}{Average Utterance Length} & 18.8  & 13.6& 16.7\\
\midrule
\multirow{5}[0]{*}{Seeker/Speaker1} & \# Utterances & 20K&53.8k& 5.7K \\
  & Avg \# Utterances & 15.4 & 4.1& 2.2 \\
   & Avg Uttr Len & 16.8& 13.2& 20.8  \\
    & \# Strategies & -& 4& -\\
    & \# Emotions & 11& 7& 32 \\
\midrule
\multirow{5}[0]{*}{Supporter/Speaker2} & \# Utterances & 18K & 49.2k& 5.2K\\
 & Avg \# Utterances & 13.6& 3.9& 2.1  \\
   & Avg Uttr Len & 21.0& 14.1& 12.3  \\
   & \# Strategies & 8& 4& -\\
  & \# Emotions & -& 7& 32 \\
\bottomrule
\end{tabular}
\caption{Statistics of ESConv, DailyDialog, and EmpatheticDialogues. 
}
\label{tab:statistics}
\end{table*}

\section{More Dataset Details}
\label{appendix:dataset_detail}

Table \ref{tab:statistics} summarizes the basic statistical information of the three datasets. The following subsections exhibit their detailed introductions.

\begin{table}[ht!]
\centering
\small
\begin{tabular}{cc|cc}
\toprule
\multicolumn{2}{c}{ESConv} &\multicolumn{2}{c}{DailyDialog}\\
\toprule
Emotion & Occurrence & Emotion & Occurrence \\
\midrule
anger & 111 & no emotion & 72143\\ 
anxiety & 354 & happiness & 11182\\ 
depression & 334 & surprise & 1600 \\ 
disgust & 40 & fear & 146 \\ 
fear & 95 & disgust & 303\\
nervousness & 13 & sadness & 969 \\ 
sadness & 308 & anger & 827 \\ 
shame & 42 \\ 
\bottomrule
\Xhline{1.2pt}
\end{tabular}
\caption{Emotion statistics of ESConv and DailyDialog.}
\label{tab:ESConv_emotions}
\end{table}



Table \ref{tab:ESConv_emotions} lists the emotion types and their occurrences in the dataset. The emotion types include anger, anxiety, depression, disgust, fear, nervousness, sadness, and shame.

\begin{table*}[h!]
\centering
\small
\resizebox{0.98\textwidth}{!}{
\begin{tabular}{>{\centering\arraybackslash}m{0.1\textwidth}m{0.25\textwidth}m{0.1\textwidth}m{0.65\textwidth}}
\Xhline{1.2pt}
\multicolumn{1}{l}{\textbf{Dataset}} &\multicolumn{1}{l}{\textbf{Strategies}} & \multicolumn{1}{c}{\textbf{Abbreviation}} &\multicolumn{1}{c}{\textbf{Definitions}}  \\ 
\toprule
\multirow{14}{*}{\textbf{ESConv}}&Question & Que.& Inquiring about problem-related information to help the seeker clarify their issues, using open-ended questions for best results and closed questions for specific details. \\ 
&Restatement or Paraphrasing &Res.\& Par.& A simple, more concise rephrasing of the help-seeker’s statements that could help them see their situation more clearly. \\ 
&Reflection of Feelings &Ref.& Articulate and describe the help-seeker’s feelings. \\ 
&Self-disclosure &Self-Dis.& Divulge similar experiences that you have had or emotions that you share with the help-seeker to express your empathy. \\ 
&Affirmation and Reassurance &Aff. \& Rea.& Affirm the help seeker’s strengths, motivation, and capabilities and provide reassurance and encouragement. \\
&Providing Suggestions &Pro.& Provide suggestions about how to change, but be careful to not overstep and tell them what to do. \\ 
&Information & Inf.& Provide useful information to the help-seeker, for example with data, facts, opinions, resources, or by answering questions. \\ 
&Others &others& Exchange pleasantries and use other support strategies that do not fall into the above categories. \\ 
\toprule
\multirow{7}{*}{\textbf{DailyDialog}}
&Inform &-& Provide factual or contextual information that the speaker believes the listener may not know or is unaware of. \\ 
&Question &-.& Seek specific information from the listener, assuming they possess the knowledge being requested. \\ 
&Directives &-& Express the speaker’s intention for the listener to take an action, including requests, instructions, or suggestions. \\ 
&Commissive &-& Indicate the speaker’s commitment to perform certain actions, such as accepting or rejecting requests or offers. \\
\Xhline{1.2pt}
\end{tabular}
}
\caption{Strategy definitions and abbreviations of ESConv and DailyDialog.}
\label{tab:ESConv_strategies}
\end{table*}

Table \ref{tab:ESConv_strategies} provides definitions of support strategies in ESConv.

\section{More Implementation Details}
\label{appendix:impelment_detail}

\subsection{Baselines}

We consider the following baselines:


\begin{itemize}
    \item \textit{Prompt-based}: including Direct (direct inference), Direct-Refine, Self-Refine \cite{Madaan2023SelfRefine}, and Chain-of-Thoughts (CoT) \cite{wei2022chain}. For CoT, we use a similar trick to the original methodology, which first generates the seeker's \textit{emotion}, which then guides the generation of strategy and response; Direct-Refine is a straightforward refinement method, in which the model revises its initial response to incorporate emotional support considerations; Self-Refine is a method that initiates by generating feedback emphasizing emotional support from the initial response, then refining the response based on this feedback \cite{Madaan2023SelfRefine}.
    \item \textit{Finetuning-based}: including {\vanillaFT} (the conventional SFT on the ESConv dataset), and CoT + SFT (first use CoT to sample from the training set of ESConv, then conduct SFT on this sampled dataset).
\end{itemize}





Table \ref{tab:prompt_baseline} shows the detailed prompts of our Direct and finetuning-based baselines. For other baselines, one can refer to \citet{kang-etal-2024-large} for prompt details.

\begin{table*}[t!]
\centering
\small
\resizebox{0.99\textwidth}{!}{
\begin{tabular}{l}
    \toprule[1.5pt]
    \begin{tabular}{l p{14cm}}
        \textbf{Methods} & \textbf{Prompt} \\ 
        \toprule
        vanilla &  
            \begin{tabular}[t]{p{14cm}}
            You will be provided with a dialogue context between a supporter and seeker. Your task is to make the next response based on the given dialogue context.\\
            The strategy should be chosen from the following 8 types of strategy: \\
            - Question:  Asking for information related to the problem to help the help-seeker articulate the issues that they face. Open-ended questions are best, and closed questions can be used to get specific information. \\
            - Restatement or Paraphrasing: A simple, more concise rephrasing of the help-seeker's statements that could help them see their situation more clearly.\\
            - Reflection of Feelings: Articulate and describe the help-seeker's feelings.\\
            - Self-disclosure: Divulge similar experiences that you have had or emotions that you share with the help-seeker to express your empathy.\\
            - Affirmation and Reassurance: Affirm the help seeker's strengths, motivation, and capabilities and provide reassurance and encouragement.\\
            - Providing Suggestions: Provide suggestions about how to change, but be careful to not overstep and tell them what to do.\\
            - Information: Provide useful information to the help-seeker, for example with data, facts, opinions, resources, or by answering questions.\\
            - Others: Exchange pleasantries and use other support strategies that do not fall into the above categories.\\
            \\
            \#\#\# Dialogue background \#\#\# \\
            The following is a conversation between a supporter and a seeker about $\{$emotion type$\}$ regarding a/an $\{$problem type$\}$. The seeker says "$\{$situation$\}$". \\
            \\
            \#\#\# Dialogue context \#\#\# \\
            $\{$context$\}$\\
            \\
            \#\#\# Model's response \#\#\#
            \end{tabular} \\
        \midrule[1pt]
        \textbf{Methods} & 
        \begin{tabular}[t]{p{9.5cm}p{4.5cm}}
                \textbf{Prompt} & \textbf{Response}
        \end{tabular} \\
    \midrule[1pt]
        {\vanillaFT} & \begin{tabular}[t]{p{9.5cm}p{4.5cm}}
                \begin{tabular}[t]{p{9.5cm}}
            <History>\\
            $\{$context$\}$\ \\
            \\
            1.Please choose the supporter's reply rule based on the above conversation history and state from the following options: \{Question, Restatement or Paraphrasing, Reflection of feelings, Self-disclosure, Affirmation and Reassurance, Providing Suggestions, Information, Others\}. \\
            2.Based on the state and rule, give a response as supporter.
        \end{tabular} & 
        \begin{tabular}[t]{p{4.5cm}}
            <Rule>\\
            $\{$strategy$\}$\ \\
            \\
            <Response>\\
            $\{$response$\}$\ \\
        \end{tabular}
        \end{tabular} \\
    \midrule
        \begin{tabular}[t]{@{}l@{}}
              Emotional-CoT + SFT \\
              (generate seeker's emotion)
        \end{tabular} & \begin{tabular}[t]{p{9.5cm}p{4.5cm}}
                    \begin{tabular}[t]{p{9.5cm}}
                <History>\\
                $\{$context$\}$\ \\
                <State>\\
                \\
            \end{tabular} & 
                \begin{tabular}[t]{p{4.5cm}}
                    <State>\\
                    $\{$state$\}$\ \\
                \end{tabular}
            \end{tabular} \\
        \begin{tabular}[t]{@{}l@{}}
              Emotional-CoT + SFT \\
              (generate supporter's strategy)
        \end{tabular} & \begin{tabular}[t]{p{9.5cm}p{4.5cm}}
                    \begin{tabular}[t]{p{9.5cm}}
                <Rule> \{Question, Restatement or Paraphrasing, Reflection of feelings, Self-disclosure, Affirmation and Reassurance, Providing Suggestions, Information, Others\}?\\
                \\
            \end{tabular} & 
                \begin{tabular}[t]{p{4.5cm}}
                    <Rule>\\
                    $\{$strategy$\}$\ \\
                \end{tabular}
            \end{tabular} \\
        \begin{tabular}[t]{@{}l@{}}
              Emotional-CoT + SFT \\
              (generate supporter's response)
        \end{tabular} & \begin{tabular}[t]{p{9.5cm}p{4.5cm}}
                    \begin{tabular}[t]{p{9.5cm}}
                <Response>\\
                \\
                \\
            \end{tabular} & 
                \begin{tabular}[t]{p{4.5cm}}
                    <Response>\\
                    $\{$response$\}$\ \\
                \end{tabular}
        \end{tabular} \\
    \end{tabular}\\
    \bottomrule[1.5pt]
\end{tabular}
}
\caption{Prompts of baselines employed for response generation.}
\label{tab:prompt_baseline}
\end{table*}

\subsection{Evaluation on Strategy Determination}

\paragraph{Proficiency.}
We define \textbf{proficiency} as \textit{how well the model selects the correct strategy}.
The proficiency for strategy ($q_{i}$) is quantified as the F1 score for strategy $i$.
To precisely analyze the model's proficiency, we utilize the \textbf{macro-F1} score, which stems from the proficiency of each strategy.
The macro F1 score ($F1$) represents the overall proficiency of the model across the strategies.


\paragraph{Preference Bias.}
We use a standard deviation of preferences $p_i$ across the strategies as the preference \textit{bias}:
\begin{equation}
\normalsize
    \mathcal{B} = \sqrt{\frac{\sum_{i=1}^{N}(p_i - \bar{p})^2}{N}}
\end{equation}
where a higher value of $bias$ indicates that the model exhibits a clear preference for both preferred and non-preferred strategies \citep{kang-etal-2024-large}.

\subsection{Evaluation on Response Quality}

\paragraph{Automatic Metrics.} \textbf{BLEU-2} (B-2) \citep{papineni2002bleu} and \textbf{Rouge-L (R-L)} \citep{lin2004rouge} are employed as the overlapping metrics between model responses and ground-truth responses.

\textbf{BLEU-2} (B-2) \citep{papineni2002bleu} first computes the geometric average of the modified $n$-gram precisions, $p_n$, using $n$-grams up to length $N$ and positive weights $w_n$ summing to one.

Next, let $c$ be the length of the prediction and $r$ be the reference length, the BP and BLEU-2 are computed as follows:

\begin{equation}
    \mathrm{BP}=\left\{\begin{array}{ll}
1 & \text { if } c>r \\
e^{(1-r / c)} & \text { if } c \leq r
\end{array} .\right.
\end{equation}

\begin{equation}
    \mathrm{BLEU}=\mathrm{BP} \cdot \exp \left(\sum_{n=1}^N w_n \log p_n\right) .
\end{equation}

\textbf{Rouge-L (R-L)} \citep{lin2004rouge} uses LCS-based F-measure to estimate the similarity between two summaries $X$ of length $m$ and $Y$ of length $n$, assuming $X$ is a reference summary sentence and $Y$ is a candidate summary sentence, as follows:

\begin{equation}
\begin{aligned}
& R_{l c s}=\frac{L C S(X, Y)}{m} \\
& P_{l c s}=\frac{L C S(X, Y)}{n} \\
& F_{l c s}=\frac{\left(1+\beta^2\right) R_{l c s} P_{l c s}}{R_{l c s}+\beta^2 P_{l c s}}
\end{aligned}
\label{rouge_l}
\end{equation}

Where $\operatorname{LCS}(X, Y)$ is the length of a longest common subsequence of $X$ and $Y$, and $\beta=P_{l c s} / R_{\text {lcs }}$ when $\partial F_{l c s} / \partial R_{l c s}=\partial F_{l c s} / \partial P_{l c s}$. In DUC, $\beta$ is set to a very big number $(\rightarrow \infty)$. Therefore, the LCS-based F-measure, i.e. Equation \ref{rouge_l}, is Rouge-L. 

\paragraph{LLM-as-a-Judge.} Table \ref{tab:prompt_comparison_score} shows the comparison evaluation prompt. We employ GPT-4o to judge the relative performance between different approaches.

\begin{table*}[t!]
\centering
\small
\resizebox{0.99\textwidth}{!}{
\begin{tabular}{p{17cm}}
    \toprule[1.5pt]
    \textbf{Prompt}\\ 
    \toprule
    
    Please act as an impartial judge and evaluate the quality of the responses provided by two AI assistants to the user questions. You should choose the assistant that follows the provided strategy more carefully and precisely to answer the user's last utterance. You should use your emotional support expertise and knowledge to judge the quality of the response considering how well the answer follows the provided strategy. Your evaluation most importantly should consider strategy adherence and then the overall quality, naturalness, consistency and coherence of the final utterance.\\
    \\
	Begin your evaluation by comparing the responses of the two assistants and provide a short explanation. Avoid any position biases and ensure that the order in which the responses were presented does not influence your decision. Do not allow the length of the responses to influence your evaluation. Do not favor certain names of the assistants. Be as objective as possible. After providing your explanation, output your final verdict by strictly following this format: "JUDGE: [[A]]" if assistant A is better, "JUDGE: [[B]]," if assistant B is better, and "JUDGE: [[C]]" for a tie.\\
    \\
	Conversation history:\\
    \\
    \{conversation\_history\}\\
    \\
    <|The Start of Assistant A's Response|>  \\
    \\
    \{assistant\_a\_resp\}  \\
    \\
    <|The End of Assistant A's Response|> \\
    \\
    <|The Start of Assistant B's Response|>  \\
    \\
    \{assistant\_b\_resp\}  \\
    \\
    <|The End of Assistant B's Response|> \\
    \bottomrule[1.5pt]
\end{tabular}
}
\caption{The Comparison Evaluation Prompt Content by GPT-4o from \cite{madani2024steeringconversationallargelanguage}.}
\label{tab:prompt_comparison_score}
\end{table*}

\paragraph{Human evaluation criteria.}

We start with the criteria proposed by \citet{kang-etal-2024-large}. The human evaluation is aimed to align with the ultimate purpose of ESC, the seeker's \textit{satisfaction}. To achieve this, the supporter's behavior can be further classified into the following criteria:

\noindent \textit{Acceptance}: Does the seeker accept without discomfort;

\noindent \textit{Effectiveness}: Is it helpful in shifting negative emotions or attitudes towards a positive direction; 

\noindent \textit{Sensitivity}: Does it take into consideration the general state of the seeker. Furthermore, to clarify the capability of LLMs to align strategy and responses, we include Alignment.

To achieve a more elaborate assessment, we consider two more dimensions addressing the generation quality:

\noindent \textit{Fluency}: the level of fluency of response.

\noindent \textit{Emotion}: the emotional intensity of response which could affect the seeker's emotional state.

We asked human annotators to rate on different dimensions. Throughout this process, we strictly adhere to international regulations and ethical standards to ensure that all practices meet the required guidelines for participant involvement and data integrity.

\begin{table*}[htbp!]
\centering
\small
\begin{tabular}{lcccccccccc}
\toprule
\textbf{Method} & $F1\uparrow$ & p-value & $bias\downarrow$ & p-value  & B-2$\uparrow$ & p-value & R-L$\uparrow$ & p-value & D-2$\uparrow$ & p-value \\ 
\midrule
    Direct & 11.9(0.9) & <1e-7 & 1.15(0.1) & <1e-7 & 3.23(0.1) & <1e-7 & 9.87(0.2) & <1e-7 & 41.2(0.5) & <1e-7 \\ 
    CoT & 9.90(0.5) & <1e-7 & 1.46(0.1) & <1e-7 & 3.47(0.1) & <1e-7 & 10.6(0.2) & <1e-7 & 37.9(0.4) & <1e-7 \\ 
    SFT & 21.0(1.5) & 0.22 & 0.56(0.1) & 0.09 & 6.37(0.3) & <1e-7 & 15.4(0.4) & <1e-7 & 48.9(0.7) & <1e-7  \\ 
    CoT+SFT & 16.4(0.5) & <1e-7 & 2.41(0.1) & <1e-7 & 5.05(0.2) & <1e-7 & 13.7(0.3) & <1e-7 & 48.8(1.4) & <1e-7 \\ 
    \ModelName &   21.2(1.3) & --- & \textbf{0.53(0.1)} & ---& \textbf{6.92(0.2)} & --- & \textbf{17.0(0.4)} & --- & \textbf{50.0(1.3)} & ---\\ 
\bottomrule
\end{tabular}
\caption{Statistical significance analysis on auto metrics. Even columns show the mean and standard deviation of each metric, while odd columns report the $p$-values of $t$-tests under the null hypothesis $H_0: Metric_X > Metric_{\ModelName}$.}
\label{tab:significance_results_auto}
\end{table*}

\begin{table*}[htbp!]

\centering
\small
\begin{tabular}{l|ccccccc}
    \toprule
    \multicolumn{1}{c|}{\multirow{2}[4]{*}{method}} & \multicolumn{7}{c}{p-values on different dimension} \\
\cmidrule{2-8}    
  & Fluency  & Emotion & Acceptance & Effectiveness & Sensitivity & Alignment &  Satisfaction \\ 
    \toprule   
    orignal dataset & \textless{}0.05 & 0.64 & 0.54 & \textless{}0.01 & \textless{}0.05 & \textless{}0.01 & \textless{}0.05 \\
    \midrule
    GPT-4o  & 0.12 & 0.95 & 0.96 & 0.27 & 0.16 & \textless{}0.01 & 0.26 \\
    GPT-4  & 0.06 & 0.83 & 0.6 & \textless{}0.01 & \textless{}0.05 & 0.14 & 0.07 \\
    Llama3-70B-inst  & \textless{}0.01 & 0.11 & \textless{}0.01 & \textless{}0.01 & \textless{}0.01 & \textless{}0.01 & \textless{}0.01 \\
    \midrule
    Llama3-8B-inst  &  \textless{}0.01 & \textless{}0.01 & \textless{}0.01 & \textless{}0.01 & \textless{}0.01 & \textless{}0.01 & \textless{}0.01 \\
    \;+ Direct-Refine & \textless{}0.01 & \textless{}0.01 & \textless{}0.01 & \textless{}0.01 & \textless{}0.01 & \textless{}0.01 & \textless{}0.01 \\
    \;+ Self-Refine & \textless{}0.01 & \textless{}0.05 & \textless{}0.01 & \textless{}0.01 & \textless{}0.01 & \textless{}0.01 & \textless{}0.01 \\
    \;+ CoT  &  \textless{}0.01 & \textless{}0.01 & \textless{}0.01 & \textless{}0.01 & \textless{}0.01 & \textless{}0.01 & \textless{}0.01 \\
    \;+ {\vanillaFT} &  \textless{}0.01 & 0.21 & \textless{}0.01 & \textless{}0.01 & \textless{}0.01 & \textless{}0.01 & \textless{}0.01 \\
    \;+ CoT+SFT  &  0.22 & 0.64 & 0.17 & \textless{}0.01 & 0.08 & \textless{}0.05 & 0.11 \\

    \bottomrule
    \end{tabular}%
\caption{Significance test of human evaluations. The table reports p-values for the hypothesis test $H_0: Metric_X > Metric_{\ModelName}$ based on Table \ref{tab:response_quaility}.}
\label{tab:significance_results_human}
\end{table*}

The detailed manual scoring criteria are as follows:
\begin{itemize}
\item Fluency:

1: The sentence is highly incoherent, making it extremely difficult to understand and failing to convey a meaningful idea.

2: The sentence has significant incoherence issues, with only parts of it making sense and struggling to form a complete thought.

3: The sentence contains some incoherence and occasional errors, but can still convey the general meaning to a certain extent.

4: The sentence is mostly fluent with only minor errors or slight awkwardness in expression, and effectively communicates the intended meaning.

5: Perfect. The sentence is completely fluent, free of any errors in grammar, punctuation, or expression, and clearly conveys the idea.

\item Emotion:

1: The emotional expression is extremely inappropriate and chaotic, not in line with the content, and may convey wrong emotions.

2: The emotional expression has obvious flaws, either too weak or exaggerated, and is disjointed from the content.

3: The emotional expression is average. It can convey basic emotions but lacks depth and has minor issues.

4: The emotional expression is good. It can effectively convey the intended emotion with an appropriate intensity and is well integrated with the content.

5: The emotional expression is excellent. It is rich, nuanced, and perfectly matches the content, capable of evoking a strong and appropriate emotional response.

\item Acceptance:

1: The response inescapably triggers emotional resistance.

2: The response is highly likely to trigger emotional resistance.

3: The response has a possibility of emotional resistance occurring.

4: The response rarely provokes emotional resistance.

5: The response has no occurrence of emotional resistance.

\item Effectiveness:

1:  The response actually worsens the seeker's emotional distress.

2: The response carries the risk of increasing stress levels, and this outcome varies depending on the individual user.

3: The response fails to alter the seeker's current emotional intensity and keeps it at the same level.

4: The response shows promise in calming the emotional intensity; however, it is overly complicated or ambiguous for the user to fully comprehend and utilize effectively.

5: The response appears to be highly effective in soothing the seeker's emotions and offers valuable and practical emotional support. 

\item Sensitivity:

1: The response renders inaccurate evaluations regarding the seeker's state.

2: The response is characterized by rash judgments, as it lacks adequate assessment and in-depth exploration of the seeker's state.

3: The response is formulated with a one-sided judgment and a limited exploration of the seeker's state.

4: The response demonstrates an understanding that only covers a part of the seeker's state.

5: The response precisely grasps the seeker's state and is appropriately tailored according to the seeker's actual situation.

\item Alignment:

1: The response is in total contradiction to the predicted strategy.

2: The response has a minor deviation from the predicted strategy.

3: There is some ambiguity between the response and the predicted strategy.

4: The response largely matches the predicted strategy, yet it contains some ambiguous elements.

5: The response effectively makes itself consistent with the predicted strategy.

\item Satisfaction:

1: The response is extremely disappointing. It doesn't answer the question at all and is of no help.

2: The response is poor. It only gives a partial answer and leaves many doubts unresolved.

3: The response is average. It meets the basic requirements but isn't particularly outstanding.

4: The response is good. It answers the question clearly and provides some useful details.

5: The response is excellent. It not only answers the question perfectly but also offers valuable additional insights.

\end{itemize}

\section{Statistical Significance of Results}
\label{sec:significance_results}

To assess the statistical significance of automatic results, we inference our model on the ESConv test set with random seeds ranging from 1 to 100, each evaluated under identical conditions. Table \ref{tab:significance_results_auto} reports the mean and std of the metrics and the p-values of the t-tests $H_0: Metric_X > Metric_{\ModelName}$ on $F1$, $bias$, and B-2. Results validate the statistical significance of {\ModelName}'s improvements.

Regarding the statistical significance of human evaluations, Table \ref{tab:significance_results_human} presents the results of the t-test $H_0: Metric_X > Metric_{\ModelName}$, indicating that our method achieves statistically significant improvements over most of the baselines.


\end{document}